\documentclass{article} 
\usepackage{iclr2026_conference,times}

\usepackage{amsmath,amsfonts,bm}

\def\eqref#1{equation~\ref{#1}}

\def\1{\bm{1}}

\DeclareMathAlphabet{\mathsfit}{\encodingdefault}{\sfdefault}{m}{sl}
\SetMathAlphabet{\mathsfit}{bold}{\encodingdefault}{\sfdefault}{bx}{n}

\usepackage{hyperref}
\usepackage{natbib}
\usepackage{url}
\usepackage{inconsolata}
\usepackage{longtable}

\usepackage{colortbl}
\usepackage{graphicx}
\usepackage{subcaption}
\usepackage{booktabs} 
\usepackage{multirow}
\usepackage{arydshln}
\usepackage{hyperref}
\usepackage{amsmath}
\usepackage{amssymb}
\usepackage{mathtools}
\usepackage{amsthm}
\usepackage{color}
\usepackage{amsfonts}
\usepackage[ruled,vlined]{algorithm2e}

\theoremstyle{plain}

\theoremstyle{definition}

\title{Agentic Reinforcement Learning with Implicit Step Rewards}

\author{
  Xiaoqian Liu\textsuperscript{1,2},
  Ke Wang\textsuperscript{2},
  Yuchuan Wu\textsuperscript{2}, 
  Fei Huang\textsuperscript{2}
  \& Yongbin Li\textsuperscript{2} \\
  \textbf{Junge Zhang\textsuperscript{3}},
  \textbf{Jianbin Jiao\textsuperscript{1}}
\\
  \textsuperscript{1}University of Chinese Academy of Sciences  \quad
 \textsuperscript{2}Tongyi Lab\\
  \textsuperscript{3}Institute of Automation, Chinese Academy of Sciences
\\
\texttt{liuxiaoqian23@mails.ucas.ac.cn} \quad
\\
\texttt{\{wk258730,shengxiu.wyc,shuide.lyb\}@alibaba-inc.com}
}

\iclrfinalcopy 
\begin{document}

\maketitle

\begin{abstract}
Large language models (LLMs) are increasingly developed as autonomous agents using reinforcement learning (agentic RL) that reason and act in interactive environments.
However, sparse and sometimes unverifiable rewards make it extremely challenging to assign credit when training LLM agents that serve as a policy.
Recent work attempts to integrate process supervision into RL but suffers from biased annotation, reward hacking, high-variance from overly fine-grained rewards or failtures when state overlap is rare.
We therefore introduce implicit step rewards for agentic RL (\textbf{iStar}), a general credit-assignment strategy that integrates seamlessly with standard RL algorithms without relying on additional rollouts or explicit step labels.
Particularly, we alternatively optimize an implicit process reward model (PRM) with the policy model to generate implicit step rewards via a trajectory-based DPO objective. Theoretical analysis shows that this learning objective produces a step-wise reward function.
Then the implicit step rewards are used to compute step-level advantages, which are combined with trajectory (or episode)-level advantages for policy updates, creating a self-reinforcing training loop.
We evaluate our method on three challenging agent benchmarks, including WebShop and VisualSokoban, as well as open-ended social interactions with unverifiable rewards in SOTOPIA. 
Crucially, \textbf{iStar} shows superior performance over frontier LLMs and strong RL baselines across domains, achieving state-of-the-art results with higher sample-efficiency and training stability.
Further analysis also demonstrates efficient exploration by \textbf{iStar} with increased rewards in both step- and episode-level while maintaining fewer steps to achieve task success.
Code will be available soon.
\end{abstract}

\section{Introduction}

LLMs are rapidly evolving from passive generators into autonomous agents that can reason, act, and adapt strategies over long horizons, including search agents~\citep{jin2025searchr,deepresearch}, mobile and web navigators~\citep{furuta2024multimodal,bai2024digirl}, software engineering assistants~\citep{yang2025qwen3,wei2025swe}, and social or embodied intelligence~\citep{liu-etal-2025-epo,lu2025vla}.
Unlike conventional RL for LLM post-training in static, single-turn tasks~\citep{ouyang2022training,shao2024deepseekmath}, training LLM agents in interactive environments faces particular challenges: 
(1) rewards are typically sparse and delayed, complicating credit assignment to intermediate actions;
(2) trajectories are long and non-Markovian in token level, with each step consisting of a chain-of-thought (CoT)~\citep{wei2022chain} and an executable action, inflating variance when credit is pushed to individual tokens;
and (3) environments and counterparts are non-stationary, open-ended and often come with unverifiable rewards (e.g, dialogues).
Consequently, trajectory-level optimization with a single outcome reward~\citep{wang2025ragen,Wei2025WebAgentR1TW} suffers from credit assignment, yielding high-variance policy learning, brittle exploration, and limited gains on agent tasks.

Recent work has attempted to solve these problems particularly through process supervision in RL. 
For example, \citet{zeng2025reinforcing,Zou2025ReasonFluxPRMTP,zhang2025rlvmr} provide denser feedback at intermediate steps but require handcrafted step labels that are costly, biased, and vulnerable to reward hacking.
Generative reward models (GRMs) (e.g, LLM-as-judge)~\citep{liu-etal-2025-epo,zha2025rl} that predict criticality or correctness for each step reduce annotation overhead but can be noisy and inconsistent across domains.
Implicit PRMs\citep{yuan2025free,cui2025process} help in single-turn tasks, but the token-level process rewards tend to be overly fine-grained in agent training, amplifying variance and destabilizing training as trajectories grow.
Other approaches~\citep{feng2025group,choudhury2025process} compute step-level advantages by grouping identical states, an assumption that fails in open-ended language environments where state overlap is rare.
Together, these limitations raise a core question for agentic RL: \textbf{How can we design a credit-assignment strategy that is label-efficient and stable, scales to multi-turn interactions, and remains robust and generalizable to (un)verifiable rewards in open-ended environments?}

To address this, we propose implicit step rewards for agentic RL (\textbf{iStar}), a general credit-assignment strategy for LLM agents. \textbf{iStar} jointly trains an implicit PRM with the policy model (the LLM agent) using trajectories collected online.
At each training step, the policy model generates rollouts, which are ranked by an outcome reward verifier (or model) to form positive–negative trajectory pairs. The implicit PRM is optimized via a trajectory-based DPO objective on these pairs.
The updated PRM then generates implicit step rewards for each action by measuring its relative preference over the previous policy snapshot. Since this reward is computed per step, it provides dense feedback to guide exploration while staying coarse enough to keep variance under control.
When training the policy model, we combine two complementary advantages: an episode-level advantage from outcome rewards and a step-level advantage from the implicit step rewards, capturing both global task success and the contribution of individual actions. 
\textbf{iStar} is compatible with various existing RL algorithms, such as GRPO~\citep{shao2024deepseekmath}, RLOO~\citep{ahmadian-etal-2024-back}, and DAPO~\citep{Yu2025DAPOAO}, without relying on annotated step labels or additional rollouts.

Our method addresses the limitations of prior work along several dimensions:
(1) \textbf{iStar} provides step credit without annotated labels through implicit rewards derived from a trajectory-based DPO objective, which is guaranteed to be a step-wise reward function learned from trajectory preferences (see Section~\ref{theory});
(2) \textbf{iStar} stabilizes training with multi-turn RL by learning implicit rewards at the step level rather than the token-level~\citep{cui2025process};
(3) \textbf{iStar} only relies on trajectory preferences that can be sourced from (un)verifiable outcome rewards even in open-ended environments, enabling unified credit assignment across different domains.

Experiments on three challenging agent benchmarks show that \textbf{iStar} achieves superior performance over frontier LLMs and strong RL baselines, achieving state-of-the-art results in WebShop and VisualSokoban. In SOTOPIA, open-ended social interactions with unverifiable rewards, \textbf{iStar} increases goal completion by up to 14\% in self-chat and 48\% when chatting with GPT-4o~\citep{4o}. Our method can also be plugged into different RL algorithms to boost their performance. Further analysis shows higher sample efficiency and training stability of our method than vanilla RL and token-level PRM~\citep{cui2025process} baselines, as well as efficient exploration with increased rewards in both step- and episode-level and fewer steps to achieve a goal. 

\section{Preliminaries}
\paragraph{\textbf{Task formulation.}}
We consider the LLM agent task as a multi-step decision-making process, where the agent interacts with the environment to achieve a long-term goal through sequential decisions given a task prompt $x\in p(X)$.
At each timestep $t$, the agent receives an observation $o_t$ from the environment and responds with a textual action $a_t \in V^L$, where $V$ denotes the token vocabulary and $L$ the maximum generation length\footnote{Note that we prompt the LLM agent to produce a reasoning process before executing an action, and $a_t$ refers to the whole sequence consisting of both the reasoning and the action at each timestep.}.
The environment then returns a scalar reward $r_t$ and transitions to the next state. Until the last timestep $T$, the full episode consists of a trajectory $\tau=\{(o_1,a_1,r_1),...,(o_T,a_T,r_T)\}$. 
However, in real-world scenarios, rewards can be sparse and delayed, such as feedback provided only at the end of a trajectory.

\paragraph{\textbf{RL for LLMs.}}
RL addresses the agent task by optimizing the LLM agent $\pi_\theta(a_t|o_{1:t},x)$, with the objective of maximizing the expected cumulative rewards during multi-turn interactions. 
Policy gradient methods are usually used, such as PPO~\citep{schulman2017proximal}, GRPO~\citep{shao2024deepseekmath}, RLOO~\citep{ahmadian-etal-2024-back}, and REINFORCE++~\citep{hu2025reinforce++}.
These RL algorithms mainly differ in the manner of estimating advantages for policy updates. For example, PPO computes advantages with a learned value function using generalized advantage estimation. GRPO and RLOO are critic-free and form relative advantages within a group of $N$ responses for the same input prompt. GRPO centers (and often normalizes) each reward by the group mean, while RLOO uses a leave-one-out mean. REINFORCE++ instead uses batch-normalized rewards as the baseline reward.

\paragraph{\textbf{Implicit reward modeling.}}
Implicit rewards have shown effectiveness in reward modeling for LLM alignment by enabling models to infer rewards without explicit labels~\citep{ethayarajh2024model,wu2025selfplay,zhang2025iterative}.
The implicit rewards are used to evaluate the quality of model outputs, such as DPO~\citep{rafailov2024direct}). Further, \citet{rafailov2024from} demonstrates that DPO can automatically learn a Q-function. Beyond the use as reward models or Q-functions, recent work uses implicit token-level rewards for test-time reranking~\citep{yuan2025free} or single-turn RL~\citep{cui2025process}: \( r_\phi(y_t) := \beta \log \frac{\pi_\phi(y_t|y_{<t})}{\pi_{\text{ref}}(y_t|y_{<t})} \), where \( \pi_\phi \) represents the token-level reward model and \( \pi_{\text{ref}} \) the reference model. $y_t$ denotes the $t$-th token in the response $y$.

\section{Method}
In this section, we first provide an overview of our method with a definition of implicit step rewards for agentic RL. We then present theoretical analysis to justify the learning objective of the implicit PRM in \textbf{iStar} produces a step-wise reward function.

\begin{figure*}[t]
    \centering
    \includegraphics[scale=0.35]{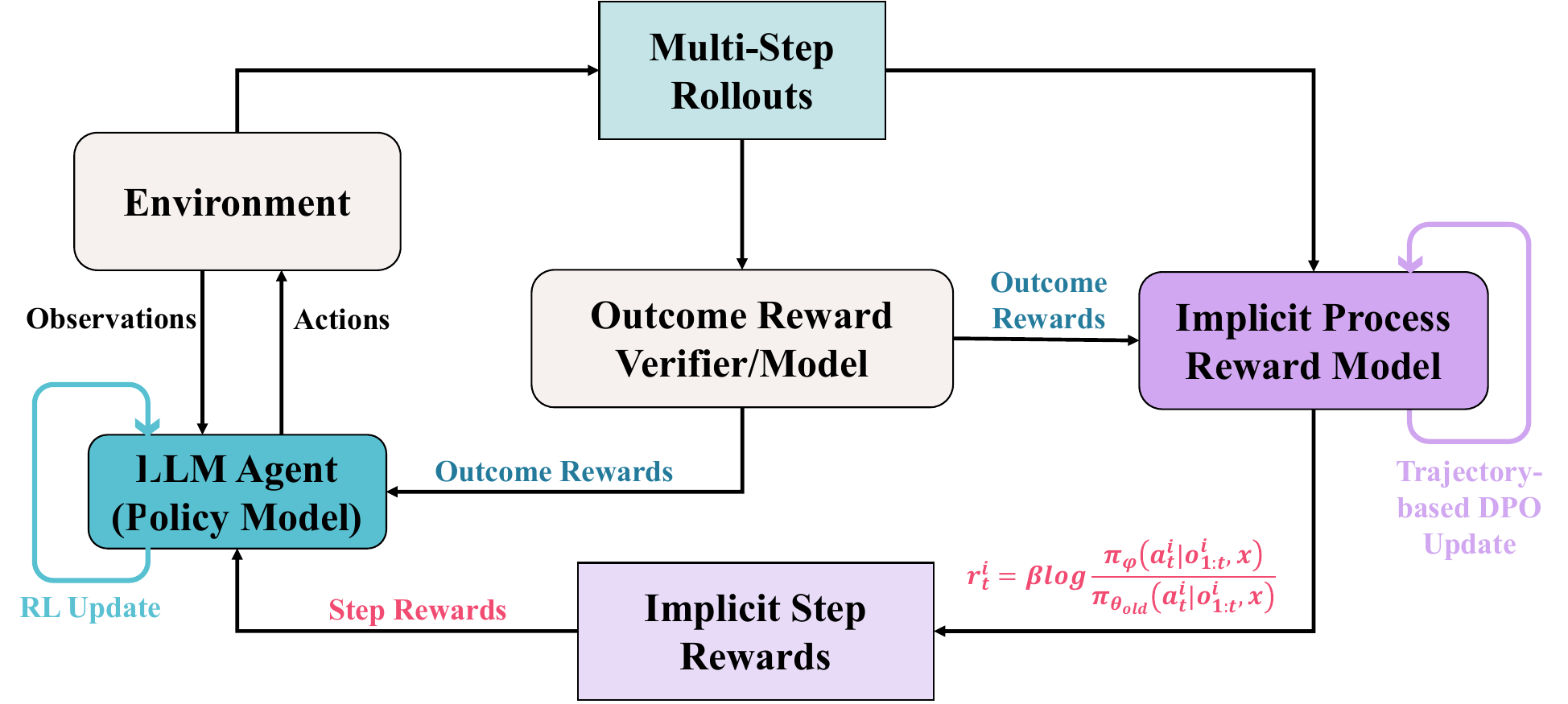}
    \caption{\textbf{Overview of iStar.} At each training step, an LLM agent interacts with the environment to generate multi‑step rollouts ranked by an outcome reward verifier (or model) to construct positive-negative trajectory pairs. These pairs are used to train an implicit PRM via a trajectory-based DPO objective, which generates implicit step rewards for each action produced by the agent.
    Finally, calculate step-level advantages using the implicit step rewards and episode-level advantages using outcome rewards to optimize the LLM agent (policy model) through RL. }
    \label{fig:method}
\end{figure*}

\subsection{Overview}

In \textbf{iStar}, there is an implicit PRM optimized alternately with the policy model (LLM agent), transforming the tendency to prefer more optimal actions into dense step rewards to guide exploration and improvement of the agent.
Figure~\ref{fig:method} shows the overall training pipeline of our method.
The alternating optimization between the implicit PRM and the policy model creates a self-reinforcing training loop to iteratively enhance each other.
Below we first provide a definition of the implicit step rewards, and detail the training process for the implicit PRM and the policy model, respectively.
Please refer to Appendix~\ref{append:algo} for the detailed algorithm of \textbf{iStar}.

\paragraph{\textbf{Implicit step rewards.}}

Let $\tau=(o_1,a_1,...,o_T,a_T)$ denote a trajectory produced by the LLM agent using policy $\pi_\theta$.
For action $a_t$ in the trajectory at step $t$, its implicit step reward is defined as
\begin{equation}
    r_\phi(o_{1:t},a_t) = \beta\log\frac{\pi_\phi(a_t|o_{1:t},x)}{\pi_{\theta_{\text{old}}}(a_t|o_{1:t},x)},
\label{eq:step_reward}
\end{equation}
where $\pi_\phi$ represents the implicit PRM, $\pi_{\theta_{\text{old}}}$ refers to the previous snapshot of the policy $\pi_\theta$, and $\beta\in[0,1]$ is a temperature that scales the reward.
The implicit step reward measures how much more probable the current action is under the freshly learned PRM than under the old policy. Positive values indicate actions that $\pi_\phi$ believes to be responsible for recent improvements, while negative values highlight actions that should be discouraged.

\paragraph{Optimizing implicit PRM via trajectory-based DPO.}
For scalable online RL, we train the implicit PRM $\pi_\phi$ on positive-negative trajectory pairs sampled by the policy $\pi_\theta$ and derive a trajectory-based DPO objective:
\begin{equation}
    \mathcal{J}_{\text{PRM}}(\phi) = -\mathbb{E}_{\substack{(\tau^{+},\tau^{-})\sim \pi_{\theta_{\text{old}}} \\ x\sim p(X)}}\left[\log\sigma\big(\beta \log\frac{\pi_\phi(\tau^{+}|x)}{\pi_{\theta_{\text{old}}}(\tau^{+}|x)}-\beta \log\frac{\pi_\phi(\tau^{-}|x)}{\pi_{\theta_{\text{old}}}(\tau^{-}|x)}\big)\right],
\label{eq:prm_loss}
\end{equation}
where $\sigma$ is the logistic sigmoid, and $\pi_\phi$, $\pi_{\theta_\text{old}}$ as well as $\beta$ follows Eq.~\ref{eq:step_reward}. 
$\tau^{+}$ is a positive trajectory that is preferred to the negative one $\tau^{-}$, both of which are labeled by an outcome reward verifier (or model) \footnote{In our experiments on WebShop and VisualSokoban, positive trajectories are those with success rates above 0, while for SOTOPIA, positive trajectories have goal completion score above 6.}. 
Particularly, Eq.~\ref{eq:prm_loss} has two main differences from the original DPO objective~\citep{rafailov2024direct}: (1) we use the previous policy snapshot $\pi_{\theta_\text{old}}$ as the reference model, whose parameters alter during training, instead of the initial policy model that keeps frozen;
(2) preferences are learned from pairwise trajectories instead of single-shot responses.
See Section~\ref{theory} for the justification that Eq.~\ref{eq:prm_loss} equivalents to a Bradley-Terry (BT) model with a step-wise reward function. 

\begin{figure*}[t]
    \centering
    \includegraphics[scale=0.3]{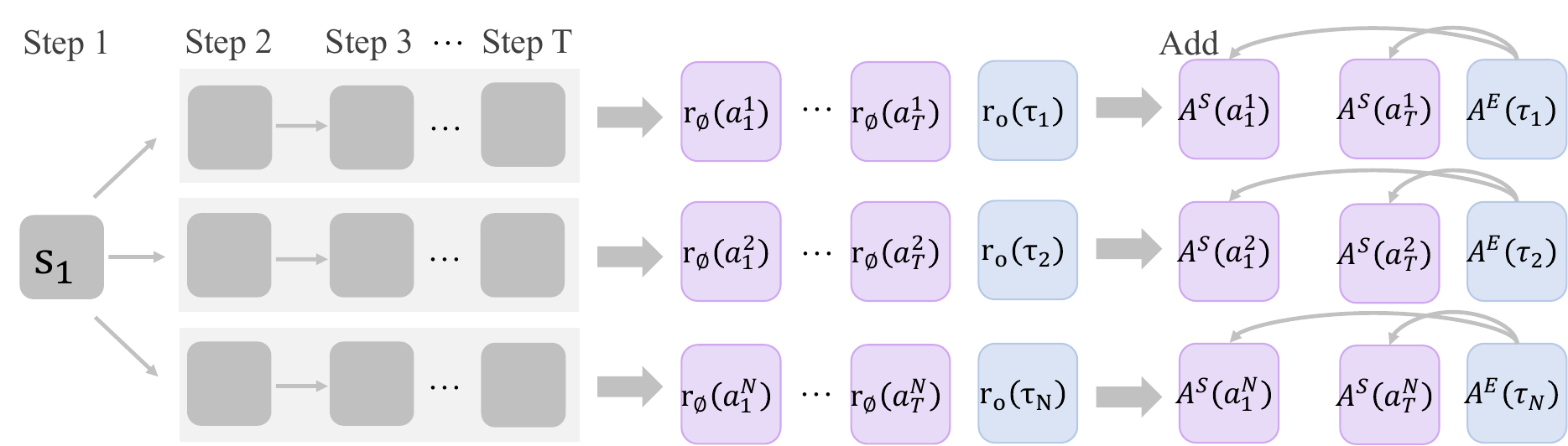}
    \caption{\textbf{The credit-assignment strategy of iStar.} In our method, episode-level advantages $A^E(\tau)$ are computed using outcome rewards $r_o(\tau)$, while step-level advantages $A^S(a)$ are calculated based on implicit step rewards $r_\phi(a)$ produced by the implicit PRM. The final advantages for policy updates is a combination of these two-level advantages.}
    \label{fig:adv}
\end{figure*}

\paragraph{\textbf{Policy learning with implicit step rewards.}}
We use GRPO~\citep{shao2024deepseekmath} as an example to illustrate how to integrate our implicit step rewards into policy learning, despite that our method is compatible with a variety of RL algorithms~\citep{ahmadian-etal-2024-back,hu2025reinforce++,Yu2025DAPOAO,Zheng2025GroupSP}.
As shown in Figure~\ref{fig:adv}, for each task prompt $x$, we sample a group of $N$ trajectories $\{\tau_1,...,\tau_N\}$ from the current policy $\pi_\theta$, and obtain its corresponding outcome rewards $\{r_o(\tau_1),...,r_o(\tau_N)\}$ through an outcome reward verifier (or model).
We then compute episode-level advantages $A^E$ for the group of trajectories:
\begin{equation}
\label{eq:episode_adv}
    A^E(\tau_i)=\big(r_o(\tau_i)-mean(R_o)\big)/std(R_o),
\end{equation}
where $R_o=\sum_{i=1}^{N}r_o(\tau_i)$. 
Next, we use the latest implicit PRM $\pi_\phi$ (from the previous training step) to obtain implicit step reward for each action $a_t^i$ via Eq.~\ref{eq:step_reward}, and compute step-level advantages:
\begin{equation}
\label{eq:step_adv}
    A^S(a_t^i)=\left(r_\phi(a_t^i)-mean(R_s)\right)/std(R_s),
\end{equation}
where $R_s = \cup_{i,t} r_\phi(a_t^i)$ denotes the whole set of step rewards in the $N$ trajectories. 
Particularly, when we use a group of trajectories starting from the same initial state (task prompt), we can generate various counterfactual scenarios. This helps us calculate a more accurate and stable estimate of the state-value baselines, leading to a better measure of the advantage for $a_t^i$. On the other hand, in a single trajectory, actions happen in different states and are influenced by noise specific to the policy used by the agent, leading to advantages in high variance.

Given episode-level advantages and step-level advantages, we combine them for policy updates:
\begin{equation}
\label{eq:overall_adv}
    A(a_t^i) = A^E(\tau_i) + \alpha{A^S(a_t^i)},
\end{equation}
where $\alpha$ is a hyperparameter that balances the two-level advantages.
The final advantage can differentiate not only between favorable and unfavorable trajectories but also beneficial and detrimental steps within a group of trajectories from the same initial state, enabling more dense rewards for policy learning in long-horizons.

Ultimately, given the advantages $A(a_t^i)$, the policy model $\pi_\theta$ is trained using a surrogate objective that is widely used in~\citet{schulman2017proximal,shao2024deepseekmath} but without KL-divergence penalty:
\begin{equation}
    \mathcal{J}_{\text{policy}}(\theta) = \mathbb{E}_{\substack{\{\tau_i\}_{i=1}^N \sim \pi_{\theta_\text{old}} \\ x \sim p(X)}} \left[
    \frac{1}{N T} \sum_{i=1}^N \sum_{t=1}^T 
    \min \left( 
        \rho_\theta(a_t^i) A(a_t^i),
         \text{clip}\left(\rho_\theta(a_t^i), 1\pm\epsilon\right) A(a_t^i) 
    \right) 
\right],
\label{eq:policy_loss}
\end{equation}
where $\rho_\theta(a_t^i) =
\frac{\pi_\theta(a_t^i | \mathbf{o}_t^i, \mathbf{x})}
{\pi_{\theta_{\text{old}}}(a_t^i | \mathbf{o}_t^i, \mathbf{x})}
$ is the importance sampling ratio at the step-level, and \( \epsilon \) is a hyperparamter that controls the clipping range of the importance.
Particularly, the step-level importance sampling ratio aligns well with our implicit step rewards to ensure low-variance training noise on multi-step rollouts, similar to~\citep{Zheng2025GroupSP}.

\paragraph{\textbf{Remarks.}}
Alternating optimization between the implicit PRM and the policy model establishes a training loop that enhances stability and accelerates convergence.
In particular, using rollouts produced by the current policy to train both models keeps their data distributions roughly consistent, minimizing off-policy bias and covariate shift. This consistency keeps implicit step rewards calibrated to the agent’s behavior, yielding dense and low-variance credit signals and preventing over- or underfitting the implicit PRM to ensure stable training.
The result is a self-reinforcing loop: improved policies yield better preference data, refining the implicit PRM, which in turn delivers more accurate implicit step rewards to guide the policy.

\subsection{Theoretical analysis}
\label{theory}

We now justify that the learning objective of our implicit PRM in Eq.~\ref{eq:prm_loss} equivalents to a BT model with a step-wise reward function.
Formally, for any trajectory pair $\{\tau_i=\{(o_t^i,a_t^i)\}_{t=1}^T\}_{i=1}^2$ satisfying $o_1^1=o_1^2$ (the same task prompt $x$), we have
\begin{equation}
\label{eq:theo_1}
\begin{split}
    & \mathbb{P}(\tau_{1}\succ \tau_{2})=\sigma\!\left(\sum_{t=1}^{T} \beta \log \frac{\pi^{*}_{\phi}(a_t^{1}\mid o_{1:t}^{1},x)}{\pi_{\theta_\mathrm{old}}(a_t^{1}\mid o_{1:t}^{1},x)}
-\sum_{t=1}^{T} \beta \log \frac{\pi^{*}_{\phi}(a_t^{2}\mid o_{1:t}^{2},x)}{\pi_{\theta_\mathrm{old}}(a_t^{2}\mid o_{1:t}^{2},x)}\right) \\
    & =\sigma\!\left(\sum_{t=1}^{T} r_\phi^*(o_{1:t}^{1},a_t^{1})-\sum_{t=1}^{T} r_\phi^*(o_{1:t}^{2},a_t^{2})\right)
\end{split}
\end{equation}
where $\pi_\phi^*$ denotes the optimal implicit PRM.
Based on the autoregressive nature of policies, Eq.~\ref{eq:theo_1} aligns with the learning objective of DPO~\citep{rafailov2024direct}, but with $\pi_{\theta_\text{old}}$ as the reference model.
Similar to the learning objective in a token-wise setup in~\citet{zhong2025dpo} that is equivalent to a BT model with a token-wise reward function \( r^*(x,y) = \beta \log \frac{\pi_\phi^*(y_t|x,y_{1:t-1})}{\pi_{\text{ref}}(y_t|x,y_{1:t-1})} \), Eq.~\ref{eq:theo_1} shows that the learning objective in Eq.~\ref{eq:prm_loss} equivalents to a BT model with a step-wise reward function:

\begin{equation}
    r_\phi^{*}(o_{1:t},a_t)=\beta \log \frac{\pi^{*}_{\phi}(a_t \mid o_{1:t},x)}{\pi_{\theta_\mathrm{old}}(a_t \mid o_{1:t},x)}.
\end{equation}
This is essentially a variant of \citet{zhong2025dpo}, where the token values within an action sequence are summed and averaged at each timestep. Please refer to \citet{zhong2025dpo} for proof details.

\section{Experiments}
We evaluate \textbf{iStar} across a variety of agentic tasks to demonstrate: (1) its effectiveness in training LLM agents for long-horizon reasoning and acting; (2) high sample efficiency and training stability given implicit step rewards; (3) improved exploration efficiency evidenced by increased rewards and fewer steps; and (4) the core components of \textbf{iStar} for credit assignment in agentic RL.
Training details can be found in Appendix~\ref{append:imple}.

\paragraph{\textbf{Benchmarks.}}
We evaluate LLM agents in three challenging environments: (1) WebShop~\citep{yao2022webshop}, a text-based web environment where the agent interacts with a HTML-based website to search, nevigate, and purchase an item given an user instruction, requiring multi-step decision making; (2) VisualSokoban~\citep{SchraderSokoban2018} with $6\times6$ size, a puzzle game where the agent has to push all boxes on targets, requiring spatial reasoning and long-term planning over both visual and textual inputs; (3) SOTOPIA~\citep{zhou2024sotopia}, an open-ended social interaction environment where the agent interacts with another LLM agent given a social scenario, role profiles and private goals, requiring reasoning over the other agent's real-time strategies. During training, we use scenarios from SOTOPIA-$\pi$~\citep{Wang2024SOTOPIAIL}. See Appendix~\ref{append:env} for more details of the environments.

\paragraph{\textbf{Baselines.}} 
We compare our method against a range of competitive baselines: (1) prompting LLMs specialized in general-purpose reasoning: GPT-5~\citep{gpt5}, Gemini-2.5-Pro~\citep{gemini}, DeepSeek-R1~\citep{guo2025deepseek}, and Claude-Sonnet-4-Thinking~\citep{claude4}; (2) vanilla RL methods that only use outcome rewards: PPO~\citep{schulman2017proximal}, GRPO~\citep{shao2024deepseekmath}, RLOO~\citep{ahmadian-etal-2024-back} and REINFORCE++~\citep{hu2025reinforce++}; (3) a recent single-turn RL method: PRIME~\citep{cui2025process} that introduces token-level process rewards for policy learning; and (4) a recent agentic RL algorithm: GiGPO~\citep{feng2025group} that computes step-level advantages via same-state grouping. 

\paragraph{\textbf{Evaluation.}} 
For WebShop and VisualSokoban, we adopt Success Rate and Score (only for WebShop) as the evaluation metrics following~\citet{feng2025group}. These metrics are computed over validation instances and select the best score for comparison. 
For SOTOPIA, we report goal completion score ranging from 0 to 10, which is evaluated by GPT-4o as a proxy for human judgement following~\citet{zhou2024sotopia}. We set the tempature to 0 for the LLM judge.  
Refer to Appendix~\ref{append:prompt} for evaluation prompts used in each environment.


\begin{table*}[t]
\caption{\textbf{Performance on WebShop and VisualSokoban}. Qwen2.5-7B-Instruct and Qwen2.5-VL-7B-Instruct serve the base models for WebShop and VisualSokoban, respectively. Note that Deepseek-R1 and PPO training do not currently support multi-modal scenarios, and PRIME is only applicable to tasks with binary outcome rewards.
Results are averaged over three random seeds. } 
\label{tab:web and soko}
\begin{center}
\begin{tabular}{lcc|c}
\toprule
\multirow{2}{*}{\textbf{Method}} & \multicolumn{2}{c}{\textbf{WebShop}}  & \multicolumn{1}{c}{\textbf{VisualSokoban}} \\
& Success & Score & Success \\
\midrule
\textit{Prompting frontier LLMs (ReAct)} & & & \\
GPT-5 & 37.5 & 66.1 & 16.6\\
Gemini-2.5-Pro & 30.5 & 38.4 & 16.0 \\
DeepSeek-R1 & 29.3 & 39.8 & - \\
Claude-Sonnet-4-Thinking & 35.2 & 62.0 & 19.1 \\
\midrule
Base Model (ReAct) & 21.5 & 47.3 & 14.1 \\ 
~~+ PPO & $78.2\pm4.5$ & $86.6\pm1.1$ & - \\
~~+ GRPO & $80.1\pm1.7$ & $89.3\pm2.8$ & $85.6\pm2.8$ \\
~~+ RLOO & $77.4\pm1.1$ & $87.6\pm4.7$ & $86.3\pm0.6$ \\
~~+ REINFORCE++ & $77.0\pm3.9$ & $85.8\pm0.1$ & $81.4\pm8.8$ \\
~~+ PRIME~\citep{cui2025process} & $81.5\pm1.8$ & $91.3\pm0.6$ & - \\
~~+ GiGPO~\citep{feng2025group} & $84.1\pm3.9$ & $91.2\pm1.5$ & $85.9\pm2.6$ \\
\rowcolor{gray!30}~~+ \textbf{RLOO w/ iStar} & $\textbf{86.5}\pm2.8$ & $\textbf{93.6}\pm1.0$ & $\textbf{91.7}\pm1.2$ \\
\bottomrule
\end{tabular}
\end{center}
\end{table*}

\subsection{Main Results}
\paragraph{\textbf{Performance on benchmarks.}}
Table~\ref{tab:web and soko} showcases \textbf{iStar}'s superior performance over baselines in WebShop and VisualSokoban, with notable gains on the latter, where RL algorithms struggle with irreversible mistakes and limited foresight. Similar trends occur with much smaller base models as demonstrated in Table~\ref{tab:small_model}.
Specifically, our method achieves state-of-the-art performance and surpasses recent multi-turn RL baseline GiGPO~\citep{feng2025group} by enabling finer-grained credit assignment, distinguishing good and bad actions with implicit rewards for each step than relying on same-state grouping.
Our method also outperforms recent single-turn RL baseline PRIME~\citep{cui2025process} that uses token-level process rewards, which provides overly fine-grained rewards that complicate policy training in multi-turn RL with high-variance (see Figure~\ref{fig:train_dynamics}(a)-(b)).
In SOTOPIA, where GiGPO and PRIME are inapplicable due to open-ended state space and unverifiable rewards, Table~\ref{tab:sotopia} shows that \textbf{iStar} still achieves state-of-the-art performance. 
Particularly, compared to vanilla RL baselines, our method improves goal completion in hard social scenarios by 14\% ($7.92\rightarrow8.06$) in self-chat, and up to 48\% ($6.68\rightarrow7.16$) increase when chatting with GPT-4o. 
This demonstrates the generalizability of our method to a wide variety of interactive environments.

\begin{table*}[t]
\caption{\textbf{Performance on Sotopia.} Self-Chat: the model being evaluated interacts with itself; GPT-4o-as-Patrner: the model interacts with GPT-4o. ``Goal'' refers to the goal completion score (0-10). ``Hard'' denotes a challenging subset of scenarios that demand for advanced reasoning, and ``All'' denotes the all set of social scenarios in SOTOPIA. Results are averaged over three random seeds.}
\label{tab:sotopia}
\begin{center}
\begin{tabular}{lcc|cc}
\toprule
\multirow{2}{*}{\textbf{Method}} & \multicolumn{2}{c}{\textbf{Self-Chat}}  & \multicolumn{2}{c}{\textbf{GPT-4o-as-Partner}} \\
& Goal (Hard) & Goal (All) & Goal (Hard) & Goal (All) \\
\midrule
\textit{Prompting frontier LLMs (ReAct)} & & & &\\
GPT-5 & 7.21 & 8.95 & \textbf{7.70} & \textbf{8.90} \\
Gemini-2.5-Pro & 6.74 & 8.27 & 7.43 & 8.41 \\
DeepSeek-R1 & 6.98 & 8.56 & 7.30 & 8.44 \\
Claude-Sonnet-4-Thinking & 6.39 & 8.64 & 7.02 & 8.62 \\
\midrule
Qwen2.5-7B-Instruct (ReAct) & 5.56 & 6.77 & 5.51 & 7.30 \\
~~+ PPO & $6.63\pm0.24$ & $8.25\pm0.09$ & $6.27\pm0.14$ & $8.07\pm0.08$ \\
~~+ GRPO & $6.97\pm0.24$ & $8.31\pm0.06$ & $6.42\pm0.31$ & $7.84\pm0.06$ \\
~~+ RLOO & $5.70\pm0.16$ & $7.13\pm0.02$ & $6.09\pm0.13$ & $7.77\pm0.03$ \\
~~+ REINFORCE++ & $6.17\pm0.30$ & $7.87\pm0.09$ & $6.38\pm0.05$ & $7.93\pm0.09$ \\
\rowcolor{gray!30}~~+ \textbf{GRPO w/ iStar} & $\textbf{7.11}\pm0.19$ & $\textbf{8.42}\pm0.03$ & $\textbf{6.76}\pm0.18$ & $\textbf{8.36}\pm0.03$ \\
\midrule
Llama3.1-8B-Instruct (ReAct) & 5.89 & 6.95 & 5.82 & 7.43 \\
~~+ PPO & $7.76\pm0.14$ & $9.05\pm0.03$ & $6.64\pm0.03$ & $8.14\pm0.01$ \\
~~+ GRPO & $7.92\pm0.08$ & $9.12\pm0.02$ & $6.68\pm0.03$ & $8.14\pm0.02$ \\
~~+ RLOO & $6.48\pm0.15$ & $8.33\pm0.03$ & $6.51\pm0.14$ & $8.02\pm0.06$ \\
~~+ REINFORCE++ & $7.84\pm0.14$ & $9.06\pm0.04$ & $6.38\pm0.23$ & $7.99\pm0.10$ \\
\rowcolor{gray!30}~~+ \textbf{GRPO w/ iStar} & $\textbf{8.06}\pm0.11$ & $\textbf{9.20}\pm0.03$ & $\textbf{7.16}\pm0.14$ & $\textbf{8.45}\pm0.03$ \\
\bottomrule
\end{tabular}
\end{center}
\end{table*}

\begin{figure}[ht]
    \centering
    \includegraphics[scale=0.34]{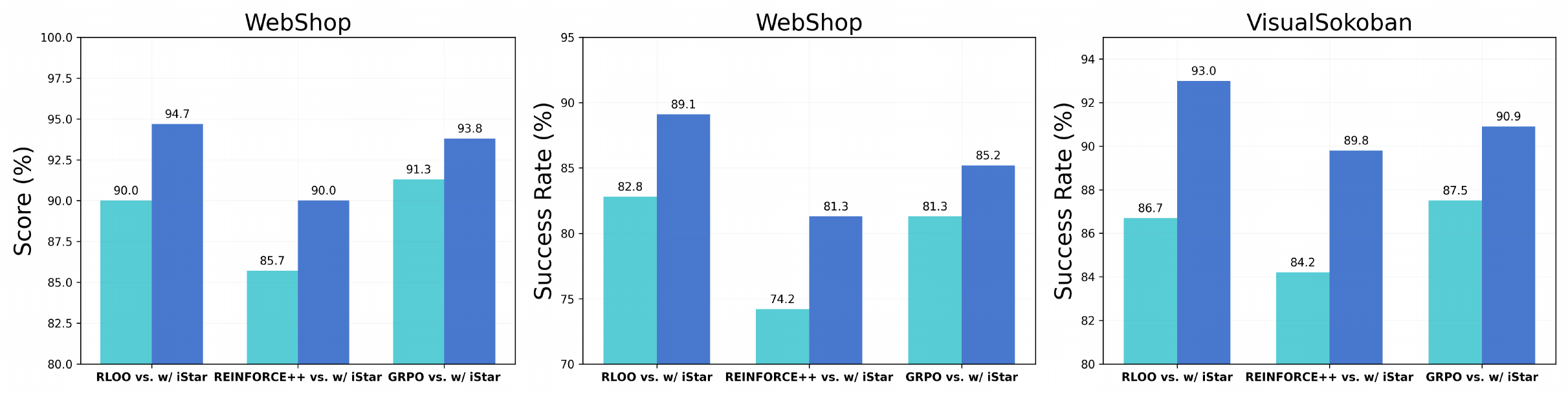}
    \caption{\textbf{Performance comparison of iStar with different vanilla RL algorithms}. Qwen2.5-7B-Instruct and Qwen2.5-VL-7B-Instruct serve the base models for WebShop and VisualSokoban, respectively. Results are reported using one seed.}
    \label{fig:all_rl_algo}
\end{figure}

\paragraph{\textbf{iStar with different vanilla RL algorithms.}}
Since our method is compatible with various RL methods, we also evaluate it using different RL algorithms, including RLOO~\citep{ahmadian-etal-2024-back}, REINFORCE++~\citep{hu2025reinforce++} and GRPO~\citep{shao2024deepseekmath}, and compare each to its vanilla version that only uses outcome rewards.
As shown in Figure~\ref{fig:all_rl_algo}, \textbf{iStar} consistently improves vanilla RL methods by integrating implicit step rewards into multi-turn RL to improve credit assignment in long-horizons.
For example, \textbf{iStar} with RLOO obtains substantial gains of 6.3\% in success rate on both WebShop and VisualSokoban. Similar trends occur with REINFORCE++ and GRPO, demonstrating the robustness of our method to diverse RL algorithms and environments.

\subsection{In-depth Analysis}
\begin{figure*}[ht]
    \centering
    \includegraphics[scale=0.3]{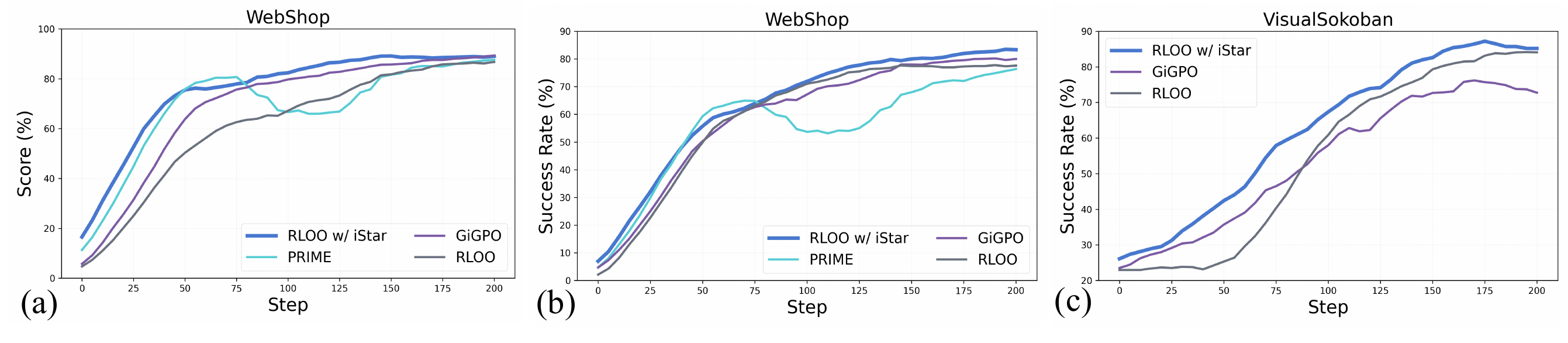}
    \caption{\textbf{Validation performance during multi-turn RL in WebShop and VisualSokoban.}
    Note that PRIME can only be applied to tasks with binary outcome rewards.}
    \label{fig:train_dynamics}
\end{figure*}

\paragraph{\textbf{Sample efficiency and training stability.}}
Figure~\ref{fig:train_dynamics} illustrates that compared to baselines, \textbf{iStar} achieves faster improvement and higher final performance in validation metrics during RL training, demonstrating superior sample efficiency.
In Figure~\ref{fig:train_dynamics}(a), our method achieves the score of vanilla RLOO in WebShop in just 105 steps, around $2\times$ improvement in training efficiency. By 165 steps, we reach the highest score of 94.7\%, showcasing significant improvements in both efficiency and performance.
Notably, while PRIME exhibits comparable early-stage performance in WebShop, its growth stagnates and experiences sharp fluctuations. This is because overly fine-grained process rewards in token-level will complicate policy learning in multi-turn interactions, which usually involve much longer sequences than single-turn tasks.
In contrast, \textbf{iStar} continues to improve consistently, suggesting that our implicit step rewards provide dense feedback for exploration while being sufficiently coarse to reduce variance for stable training.

\begin{figure*}[ht]
    \centering
    \includegraphics[scale=0.31]{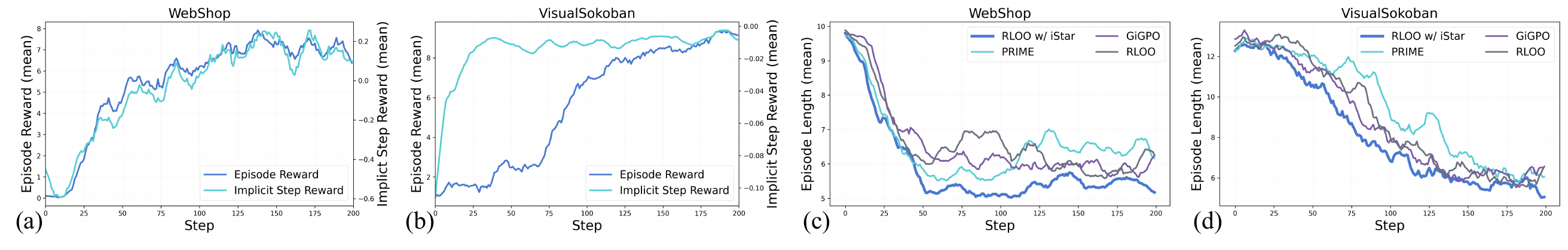}
    \caption{\textbf{Training dynamics of iStar in WebShop and VisualSokoban}. \textbf{Left:} Dynamics of the episode and implicit step rewards during RL training by our method. \textbf{Right:} The episode length versus training step compared to baselines.}
    \label{fig:explore}
\end{figure*}

\paragraph{\textbf{Exploration efficiency.}}
To demonstrate the implicit step rewards provide useful guidance for policy learning, we visualize the dynamics of both step- and episode-level rewards in Figure~\ref{fig:explore}(a) and (b). In particular, the implicit step reward improves very early (especially in VisualSokoban) and then the episode reward follows, indicating that our method first captures good local action heuristics and then composes them into higher-return trajectories.
As a result, it also reduces unnecessary actions during multi-turn interactions, leading to shorter episode lengths. As shown in Figure~\ref{fig:explore}(c) and (d), episode lengths decrease without compromising task success, as evidenced by the consistent increase in episode rewards. 

\begin{table*}[ht]
\caption{\textbf{Ablation studies on core components of iStar}. 
``RLOO'': only outcome rewards are used to compute advantages for policy updates.
``w/ ground-truth process rewards": use raw step rewards provided by VisualSokoban to calculate step-level advantages.
``w/ merged rewards'': implicit step rewards are added directly to outcome rewards before advantage computation.
``w/ token-level process rewards": implicit step rewards are computed in token level rather than step level.
Results are reported using one seed.}
\label{tab:ablation}
\begin{center}
\begin{tabular}{lcc|c}
\toprule
\multirow{2}{*}{\textbf{Method}} & \multicolumn{2}{c}{\textbf{WebShop}} & \textbf{VisualSokoban} \\
& Success & Score & Success\\
\midrule
RLOO & 76.6 & 84.2 & 85.9 \\
~~w/ ground-truth process rewards & - & - & 87.5 \\
~~w/ merged rewards & 81.3 & 90.7 & 88.3 \\
~~w/ token-level process rewards & 82.0 & 90.0 & 89.1 \\
\rowcolor{gray!30}~~\textbf{w/ iStar} & \textbf{89.1} & \textbf{94.7} & \textbf{93.0}\\
\bottomrule
\end{tabular}
\end{center}
\end{table*}

\subsection{Ablation Studies}
Table~\ref{tab:ablation} presents results of ablation experiments to validate the necessity of key components of our method for effective credit assignment.
First, raw step penalties provided in VisualSokoban show limited improvement over vanilla RL, suggesting that the implicit step rewards learned by \textbf{iStar} are superior credit signals for policy learning.
Second, merging the implicit step rewards into episode rewards obtains gains over vanilla RL but the improvement are modest compared to our method. This indicates that we should not only reward intermediate actions but also gate credit by final task success to prevent speculative reward exploitation. Therefore, combining signals at the advantage level is crucial to credit assignment in long-horizons.
Third, learning token-level process rewards is sub-optimal to multi-turn RL, suggesting that overly fine-grained rewards may introduce noise and thus increase difficulty of policy learning.

\section{Related Work}


PRMs has been widely explored in single-turn tasks, such as mathematical reasoning and single-shot code generation. In these settings, PRMs are used to score intermediate steps for test-time search or reranking~\citep{lightman2024lets,wang-etal-2024-math,Mahan2024GenerativeRM}, or for online-RL~\citep{dou-etal-2024-stepcoder,setlur2025rewarding,zha2025rl}. Moving to dynamic, interactive tasks, process rewards are usually constructed in three ways: (1) handcrafted step rewards assigned to tool execution~\citep{zeng2025reinforcing} or meta-reasoning tags~\citep{zhang2025rlvmr}; (2) GRMs~\citep{liu-etal-2025-epo,Zou2025ReasonFluxPRMTP} that label step quality; (3) and implicit PRMs~\citep{yuan2025free} that produce token-level process rewards.
However, manually-designed or judge-based step labels are costly and biased, suffering from reward hacking. Learning step Q-values~\citep{choudhury2025process} can reduce the bias but a fixed PRM may poorly estimate Q-values for unseen actions during inference.
PRIME\citep{cui2025process} partially addresses these issues by jointly training an implicit PRM with the generator. However, with overly fine-grained rewards in token-level, it introduces noises and distabilizes training in multi-turn RL. Another issue with PRIME is that it applies a cross-entropy loss to optimize the implicit PRM, which is only applicable to tasks with binary outcome rewards. 
Instead of learning a PRM, \citet{feng2025group} addresses credit assignment by computing step-level advantages via same-state grouping. While effective in tasks with finite state-action space, it relies on exact state overlaps and cannot be generalized to open-ended language environments where same state is rare.
In contrast, our method learns implicit rewards at the step-level through a trajectory-level DPO objective, offering label-efficient step rewards with low-variance and showing robustness and generalization to (un)verifiable rewards even in open-ended environments.

\section{Conclusion}
We propose \textbf{iStar}, a general credit-assignment strategy for LLM agents.
In particular, we alternatively optimize an implicit PRM with the policy model, which provides implicit step rewards to guide policy learning and results in a self-reinforcing training loop.
to optimize the implicit PRM, we propose a trajectory-level DPO objective that is guaranteed to be equivalent to a BT model with a step-wise reward function. 
In practice, our method can be plugged into a variety of RL algorithms and generalized to (un)verifiable rewards even in open-ended environments.
Empirical results show that \textbf{iStar} achieves averaged 86.5\% success and 93.6\% score on WebShop, and reaches 91.7\% success on VisualSokoban. In SOTOPIA, our method improves goal completion in hard scenarios by up to 14\% in self-chat and 48\% against GPT-4o.
In the future, our method can also be validated in math problems or code generation to provide implicit step rewards for intermediate CoT steps. It can also be applied to test-time scaling for search guidance.
Regarding the limitations, we currently separate the implcit PRM from the policy model during training, which however, can be a unified model trained with different objectives to reduce computation memory and potentially improve representation sharing.
Additionally, in SOTOPIA, our implicit PRM is trained to only predict goal-completion preferences, while future work could be extended to multi-objective implicit PRMs.

\bibliography{iclr2026_conference}
\bibliographystyle{iclr2026_conference}

\appendix
\clearpage

\section{Algorithm}
\label{append:algo}
The algorithm flow of iStar is detailed in Algorithm~\ref{algo}.
\begin{algorithm}[ht]
\caption{Training LLM Agents with iStar (GRPO as an example)}
\KwIn{Task distribution $p(X)$, language model $\pi_{\theta_{\text{init}}}$, outcome reward verifier or model $r_o$, training steps M, rollout size N, mixing weight $\alpha$}
\KwOut{Optimized policy $\pi_\theta$ and PRM $\pi_\phi$}
Initialize policy model $\pi_\theta\leftarrow{\pi_{\theta_{\text{init}}}}$, $\pi_{\theta_{old}}\leftarrow{\pi_{\theta_{\text{init}}}}$, PRM $\pi_\phi\leftarrow{\pi_{\theta_{\text{init}}}}$;\\
\For{\textnormal{iteration $= 1, ..., M$}}{
    {\color{blue}\tcp{Multi-step rollouts collection}}
    
    Sample task $x\sim p(X)$ and initialize N identical environments
    
    \For{\textnormal{$t = 1,...,T$}}{
       Sample actions $\{a_t^i\sim \pi_\theta(o_{1:t}^i,x)\}_{i=1}^N$
       
       Execute actions and observe next observation $\{o_{t+1}^i\}_{i=1}^N$}

    {\color{blue}\tcp{PRM training}}
    
    Compute outcome rewards for N trajectories: $r_o(\tau_{1:N})$
    
    Forward pass $\pi_\phi$ based on trajectory preferences to obtain step reward $r_\phi(a_t^i)$ with Eq.~\ref{eq:step_reward}
    
    Update PRM $\pi_\phi$ on trajectories using a DPO-style objective in Eq.~\ref{eq:prm_loss}

    {\color{blue}\tcp{Policy training}}
    
    Compute episode-level advantages $A^E(\tau_i)$ using $r_o(\tau_i)$ via Eq.~\ref{eq:episode_adv}
    
    Compute step-level advantages $A^S(a_t^i)$ using $r_\phi(a_t^i)$ via Eq.~\ref{eq:step_adv}
    
    Combine advantages: $A(a_t^i) = A^E(\tau_i) + \alpha{A^S(a_t^i)}$
    
    Update policy $\pi_\theta$ by maximizing objective in Eq.~\ref{eq:policy_loss}
    
    Update old parameters: $\theta_{old}\leftarrow{\theta}$}
\label{algo}
\end{algorithm}



\section{Environment Details}
\label{append:env}

\paragraph{\textbf{WebShop.}}
WebShop~\citep{yao2022webshop} simulates an online shopping task on an e-commerce platform, where the agent's objective is to interpret human-provided text instructions and purchase a product that aligns with the given specifications. To accomplish this, the agent must interact with the website's search engine, select items to review from the search results, examine their descriptions and details, and choose relevant options (e.g., size, color) before finalizing the purchase by clicking the ``Buy'' button. To identify the best product that fulfills the user’s requirements, the agent may need to compare multiple products, navigate back and forth between pages, and conduct additional searches if necessary. The environment includes over one million products sourced from amazon.com, more than 12,000 crowd-sourced instructions, and a rich set of semantic actions. Rewards are automatically calculated using programmatic matching functions that evaluate attributes, type, options, and price of the selected product.

\paragraph{\textbf{VisualSokoban.}}
Sokoban~\citep{SchraderSokoban2018} consists of rooms composed of five key elements: walls, floors, boxes, box targets, and an agent. These elements may exist in different states depending on whether they overlap with a box target. Rooms are randomly generated, which helps prevent models from overfitting to specific predefined layouts. The game includes two primary actions, Push and Move, which can be performed in four directions: Up, Down, Left, and Right. The Move action allows the agent to proceed to an empty space in the specified direction, provided there is no wall or box blocking the path. The Push action attempts to move an adjacent box, but only if the field behind the box is empty; chain pushing of multiple boxes is not allowed. If no box is adjacent, the Push action functions identically to the Move action in the same direction. Successfully completing the game by pushing all boxes onto their targets yields a reward of 10 points on the final step. Additionally, pushing a box onto a target grants a reward of 1 point, while removing a box from a target results in a penalty of -1 point. Each step incurs a small penalty of -0.1 points to discourage trajectories with many steps. VisualSokoban renders visuals in RGB, with the pixel size equal to the grid size.

\paragraph{\textbf{SOTOPIA.}}
SOTOPIA~\citep{zhou2024sotopia} is a general-domain, open-ended platform to simulate social interactions between LLM agents. The scenarios span a diverse array of social interaction types, such as negotiation, exchange, collaboration, competition, accommodation, and persuasion. A particularly challenging subset, known as SOTOPIA-hard, involves scenarios requiring advanced strategic reasoning. Each agent is defined by character profiles, encompassing attributes like name, gender, personality, and profession. At the end of each dialogue, agents are evaluated by GPT-4o across seven dimensions: Goal Completion, Believability, Knowledge, Secret, Relationship, Social Rules, Financial and Material Benefits.
SOTOPIA-$\pi$~\citep{Wang2024SOTOPIAIL} is a follow-up work that uses GPT-4 to generate a new set of scenarios. The social tasks in SOTOPIA-$\pi$ are guaranteed to be entirely distinct from those in SOTOPIA.

\section{Experiment Details}
\subsection{Training Details}
\label{append:imple}

In \textbf{iStar}, the implicit PRM is initialized from the base policy model, except for VisualSokoban, where the policy model uses Qwen2.5-VL-7B-Instruct and the PRM uses Qwen2.5-7B-Instruct. 
We use a constant learning rate $5\times10^{-7}$ for the policy model and $10^{-6}$ for the implicit PRM with AdamW optimizer. Both the policy and the implicit PRM use a batch size of 64 and micro-batch size 8. No KL-divergence penalty is applied.
We set the advantage coefficient $\alpha$ to 1 and $\beta=0.05$ for the implicit PRM training. 
The rollout size is set to 8 per prompt.
All methods share the identical RL configurations for fair comparison, and all experiments are run on $8\times$A100 GPUs.
Below we show additional training details in each environment.

\paragraph{\textbf{WebShop and VisualSokoban.}}
We use Qwen2.5-(VL)-7B-Instruct~\citep{Yang2024Qwen25TR} as the base models for policy learning.
To address invalid actions produced by LLM agents, a reward penalty of -0.1 is imposed.
The maximum response length is 512 tokens, while the maximum prompt length is 4096 tokens in WebShop and 1024 tokens in VisualSokoban.
We sample 16 different groups per rollout in WebShop, resulting in a total of $16\times8=128$ environments. In VisualSokoban, we sample 32 different groups per rollout, resulting in a total of $32\times8=256$ environments. 
Instead, PPO uses 128 and 256 separate environments for rollouts in WebShop and VisualSokoban, respectively. The rollout temperature is set to 1.0, while the validation temperature is set to 0.4.
We implement experiments in veRL~\citep{sheng2024hybridflow}, each for 200 training steps.

\paragraph{\textbf{SOTOPIA.}}
We use Qwen2.5-7B-Instruct~\citep{Yang2024Qwen25TR} and Llama3.1-8B-Instruct~\citep{llama-3.1} as the base models for policy learning to demonstrate the robustness of our method to different model backbones. 
The maximum prompt length is 6144 tokens and the maximum response length is 2048 tokens.
As with WebShop, we sample 16 different groups per rollout in WebShop, resulting in a total of $16\times8=128$ environments (PPO uses 128 separate environments).
The rollout temparature is set to 0.7.
Each experiment implemented in veRL consists of 800 training steps.

\subsection{Evaluation Prompts}
\label{append:prompt}
\begin{figure}[h]
    \centering
    \includegraphics[scale=0.4]{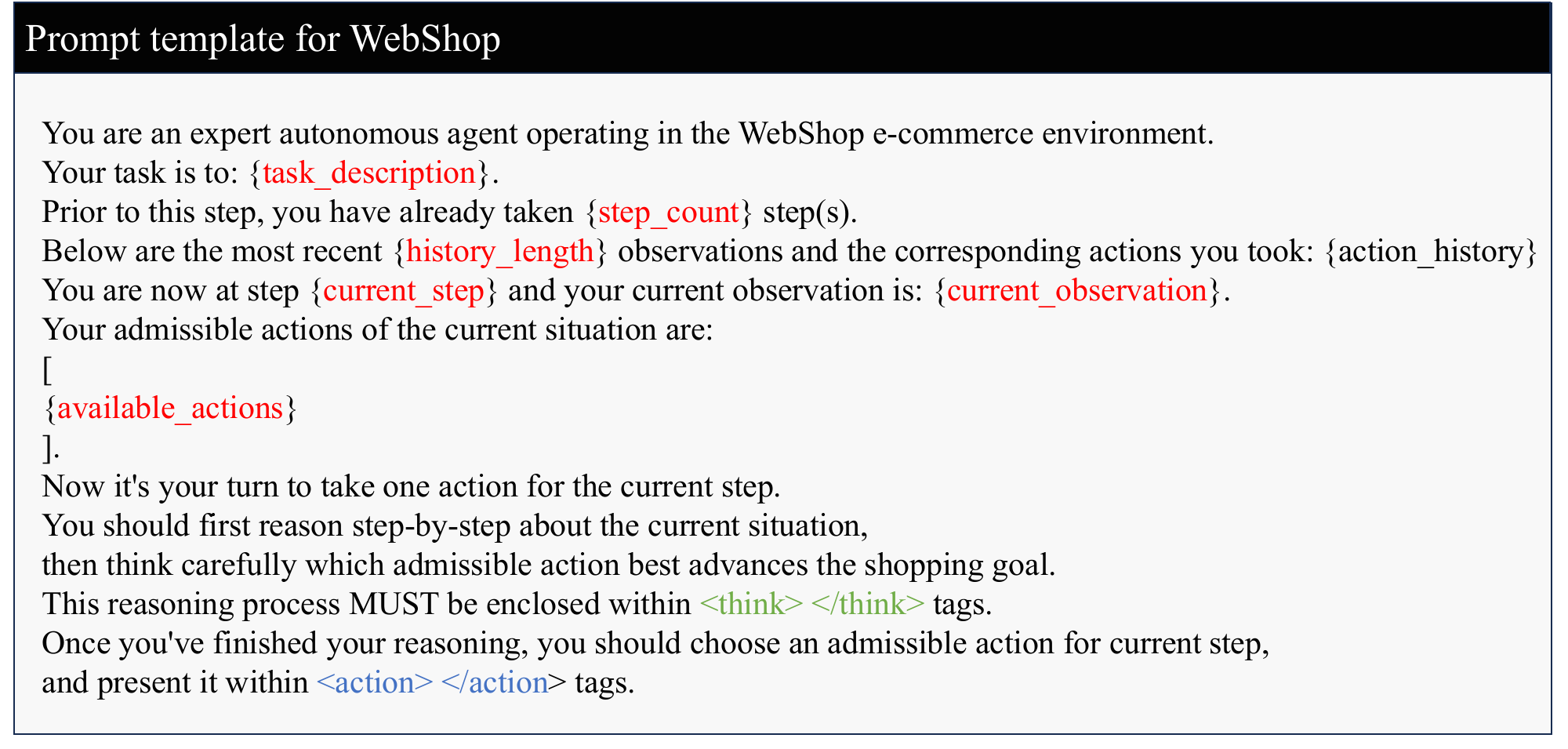}
    \caption{Evaluation prompts used for the LLM agent in WebShop.}
    \label{fig:web_prompt}
\end{figure}

\begin{figure}[h]
    \centering
    \includegraphics[scale=0.4]{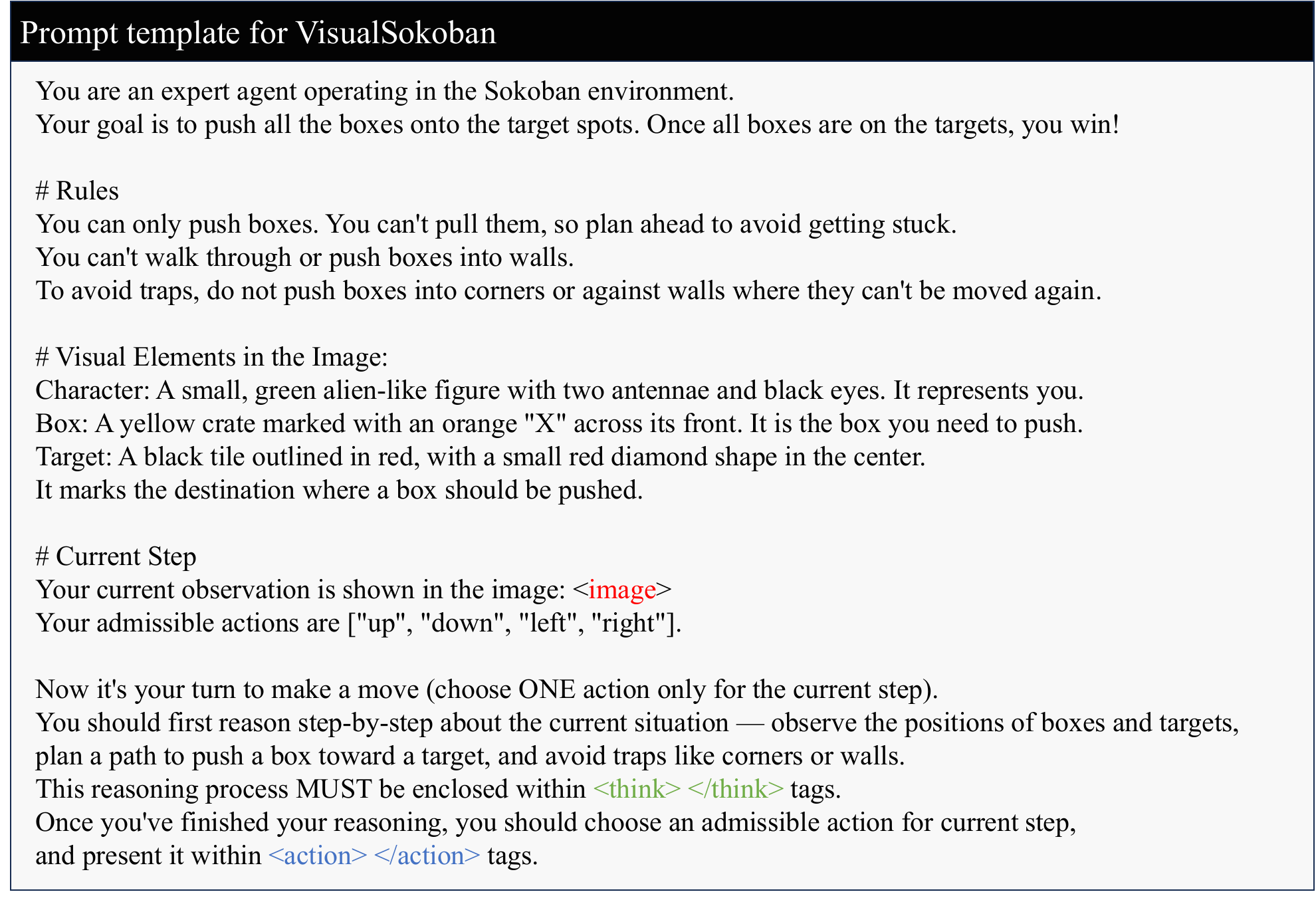}
    \caption{Evaluation prompts used for the MLLM agent in VisualSokoban.}
    \label{fig:soko_prompt}
\end{figure}

\begin{figure}[h]
    \centering
    \includegraphics[scale=0.4]{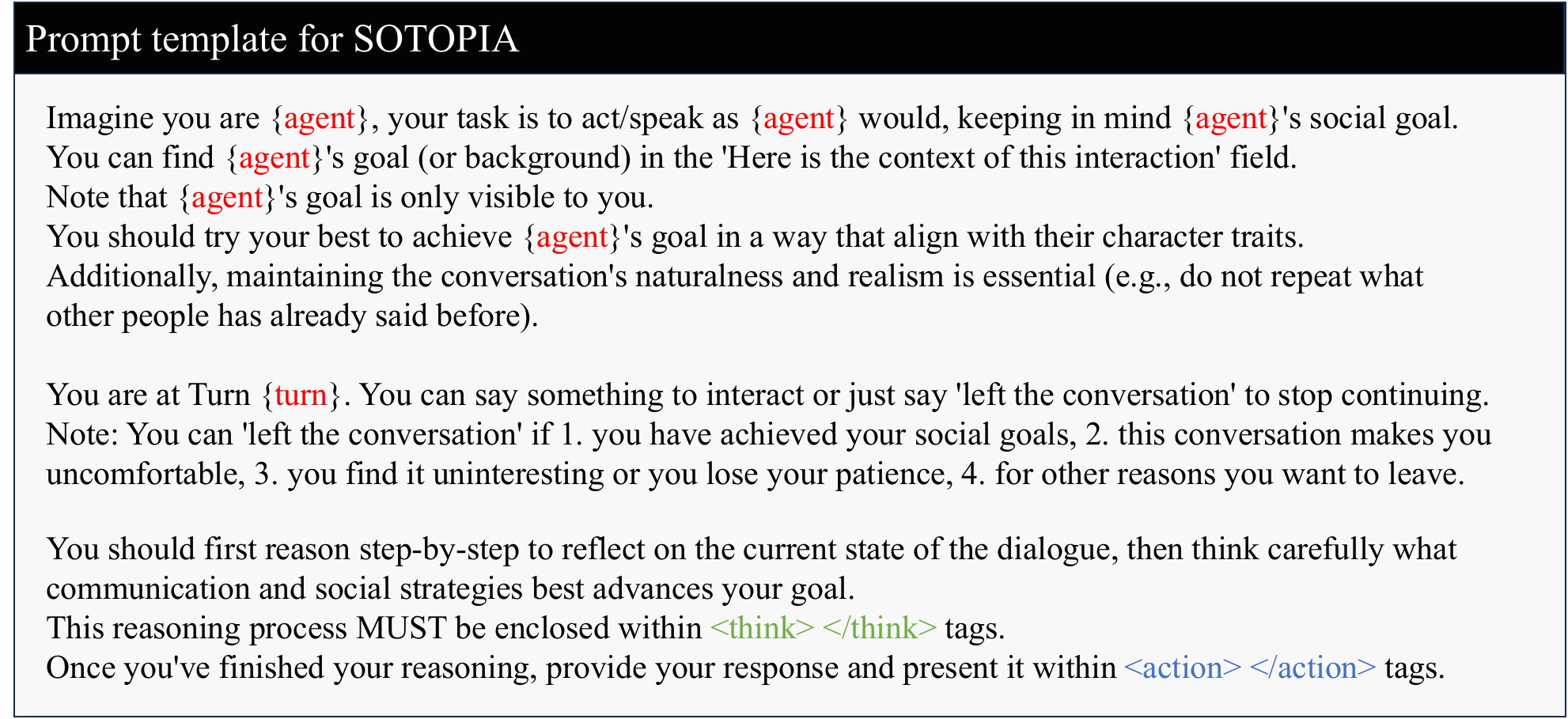}
    \caption{Evaluation prompts used for LLM agents in SOTOPIA.}
    \label{fig:soto_prompt}
\end{figure}

We use ReAct~\citep{yao2023react} as the prompting strategy, with chain-of-thought~\citep{wei2022chain} generated before each action. 
The prompt templates used for evaluating LLM or multimodal large language model (MLLM) agents in WebShop~\citep{yao2022webshop}, VisualSokoban~\citep{SchraderSokoban2018}, and SOTOPIA~\citep{zhou2024sotopia} are presented in Figure~\ref{fig:web_prompt}, Figure~\ref{fig:soko_prompt}, and Figure~\ref{fig:soto_prompt}.
Placeholders enclosed in curly braces(\{\}) represent semantic slots, which are dynamically populated at runtime.

\section{Additional Results}
\label{append:res}

Table~\ref{tab:small_model} presents results of iStar compared to baselines on WebShop using Qwen2.5-1.5B-Instruct~\citep{Yang2024Qwen25TR} as the base model.

\begin{table*}[ht]
\caption{\textbf{Performance on WebShop with Qwen2.5-1.5B-Instruct as the base model.} 
Results are reported using one seed.}
\label{tab:small_model}
\begin{center}
\begin{tabular}{lcc}
\toprule
\textbf{Method} &  \textbf{Success} & \textbf{Score} \\
\midrule
Qwen2.5-1.5B-Instruct (ReAct) & 10.4 & 46.1 \\
~~+ RLOO & 71.9 & 85.7 \\
~~+ GiGPO & 72.7 & 86.8 \\
~~+ PRIME & 74.2 & 86.6\\
\rowcolor{gray!30}~~+ \textbf{RLOO w/ iStar} & \textbf{80.5} & \textbf{91.5} \\
\bottomrule
\end{tabular}
\end{center}
\end{table*}


\end{document}